\documentclass[sigconf]{acmart}
\AtBeginDocument{%
  \providecommand\BibTeX{{%
    \normalfont B\kern-0.5em{\scshape i\kern-0.25em b}\kern-0.8em\TeX}}}

\setcopyright{acmcopyright}
\copyrightyear{2022}
\acmYear{2022}

\def\etal{\emph{et al.}}
\usepackage{mathrsfs}
\usepackage{algorithm}
\usepackage{algorithmic}
\usepackage{amsmath}  
\usepackage{color}
\usepackage{fancyhdr}
\usepackage{multirow} 
\renewcommand\footnotetextcopyrightpermission[1]{}

\begin{document}
\begin{sloppypar}
\title{Playing Lottery Tickets in Style Transfer Models}


\author{Meihao Kong}
\affiliation{%
  \city{Nanjing University}
  \country{China}
}
\email{kongmeihao@smail.nju.edu.cn}

\author{Jing Huo}
\affiliation{%
  \city{Nanjing University}
  \country{China}
}
\email{huojing@nju.edu.cn}

\author{Wenbin Li}
\affiliation{%
  \city{Nanjing University}
  \country{China}
}
\email{liwenbin@nju.edu.cn}

\author{Jing Wu}
\affiliation{%
  \city{Cardiff University}
  \country{UK}
}
\email{WuJ11@cardiff.ac.uk}

\author{Yu-Kun Lai}
\affiliation{%
  \city{Cardiff University}
  \country{UK}
}
\email{LaiY4@cardiff.ac.uk}

\author{Yang Gao}
\affiliation{%
  \city{Nanjing University}
  \country{China}
}
\email{gaoy@nju.edu.cn}

\begin{abstract}
    Style transfer has achieved great success and attracted a wide range of attention from both academic and industrial communities due to its flexible application scenarios. However, the dependence on a pretty large VGG-based autoencoder leads to existing style transfer models having high parameter complexities, which limits their applications on resource-constrained devices. Compared with many other tasks, the compression of style transfer models has been less explored. Recently, the lottery ticket hypothesis (LTH) has shown great potential in finding extremely sparse matching subnetworks which can achieve on par or even better performance than the original full networks when trained in isolation. In this work, we for the first time perform an empirical study to verify whether such trainable matching subnetworks also exist in style transfer models. Specifically, we take two most popular style transfer models, \emph{i.e.}, AdaIN and SANet, as the main testbeds, which represent global and local transformation based style transfer methods respectively. We carry out extensive experiments and comprehensive analysis, and draw the following conclusions. ($\romannumeral1$) Compared with fixing the VGG encoder, style transfer models can benefit more from training the whole network together. ($\romannumeral2$) Using iterative magnitude pruning, we find the matching subnetworks at $89.2\%$ sparsity in AdaIN and $73.7\%$ sparsity in SANet, which demonstrates that \textit{Style transfer models can play lottery tickets too}. ($\romannumeral3$) The feature transformation module should also be pruned to obtain a much sparser model without affecting the existence and quality of the matching subnetworks. ($\romannumeral4$) Besides AdaIN and SANet, other models such as LST, MANet, AdaAttN and MCCNet can also play lottery tickets, which shows that LTH can be generalized to various style transfer models.
\end{abstract}

\ccsdesc[500]{Computing methodologies~Computational photography}

\ccsdesc[500]{Computing methodologies~Non-photorealistic rendering}

\keywords{style transfer, neural network pruning, lottery ticket hypothesis}


\maketitle

\section{Introduction}
Recent years have witnessed rapid development in the area of neural style transfer, which aims at composing a content image with new styles from reference images. 
Extensive research has focused on improving visual quality~\cite{SANet, kolkin2019style, LST, liu2021adaattn,chandran2021adaptive, chen2021dualast}, efficiency~\cite{coust, johnson2016perceptual, li2016precomputed, radford2015unsupervised, ulyanov2016texture, jing2020dynamic} and flexibility~\cite{gatys2017controlling, an2021artflow, risser2017stable, hu2020aesthetic, huo2021manifold, hong2021domain}. Although great success has been achieved in these aspects, the memory and computational footprints required for these style transfer models are large owing to the widely adopted autoencoder-based architecture. In particular, they usually consist of a pre-trained VGG encoder~\cite{simonyan2014very}, a feature transformation module and a corresponding decoder, with a large number of parameters, which makes these models infeasible to be used in resource-constrained scenarios. This naturally raises a question: \textit{``Can we prune a large style transfer model while preserving its performance?''}

\begin{figure}[tp]
  \centering
  \includegraphics[width=\linewidth]{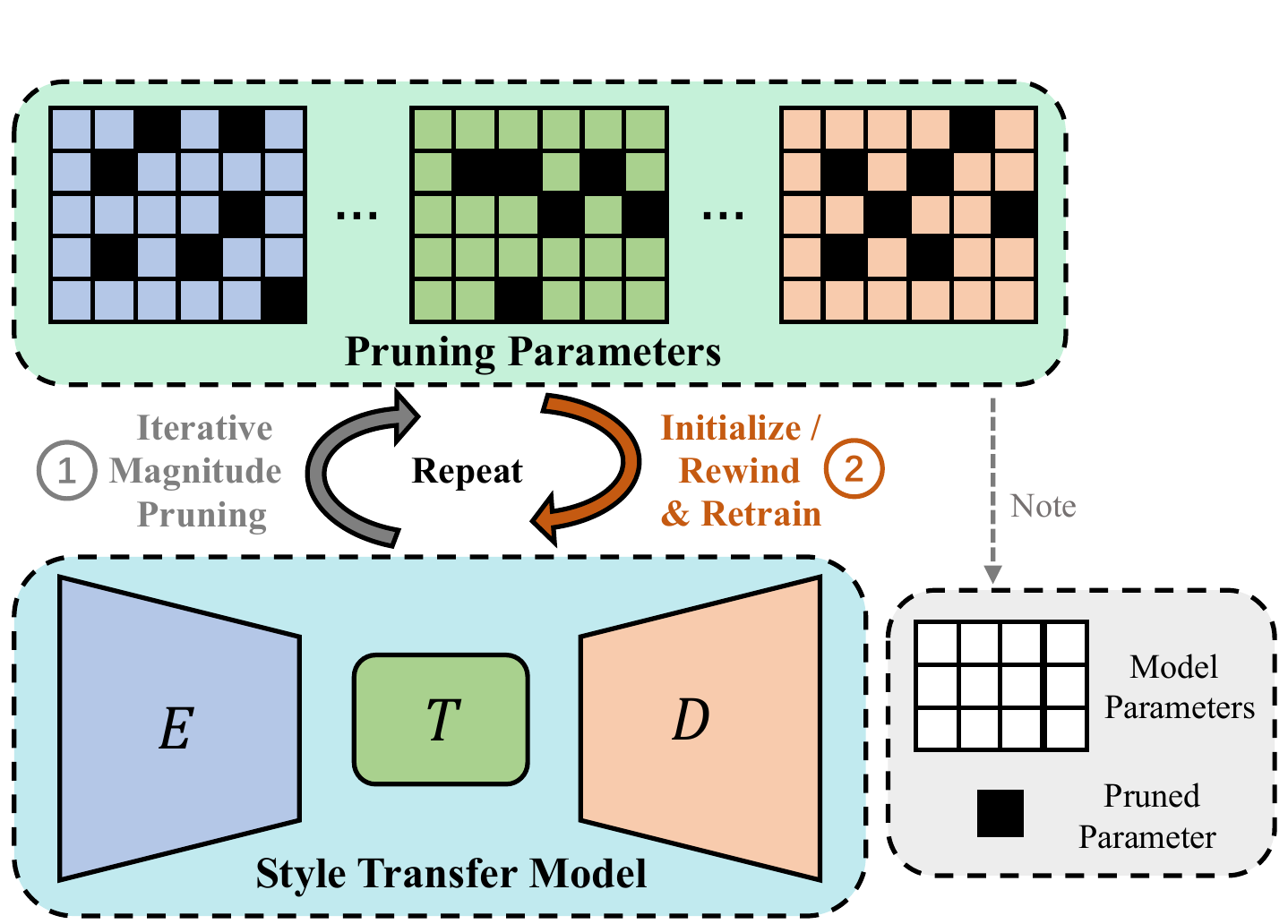}
  \caption{Overview of the overall process for playing lottery tickets in style transfer models. Winning tickets can be found by Iterative Magnitude Pruning (IMP). After that, we initialize or rewind the parameters of the original full model and then retrain the subnetwork to verify the performance of the found ticket.}
  \label{fig:Framework}
\end{figure}

In this paper, we aim to answer the above question via the lens of lottery ticket hypothesis (LTH)~\cite{frankle2018lottery}. LTH states that there exist small subnetworks in dense neural networks that can be trained in isolation from initialization to match the performance of the original network after training for at most the same number of iterations. Existence of LTH has been successfully shown in various fields~\cite{yu2019playing, renda2020comparing, chen2020lottery, gan2021playing}, and its property has been widely studied~\cite{malach2020proving, frankle2020linear, chen2021unified, kalibhat2020winning}. Nonetheless, to the best of our knowledge, no prior work exists on understanding the lottery ticket hypothesis in style transfer, which otherwise could be a powerful tool to understand the parameter redundancy in the current prevailing style transfer models. This will be the focus of our work. 

Specifically, we mainly investigate the existence of LTH in two representative style transfer models -- AdaIN (a representative \textit{global transformation based} method)~\cite{AdaIN} and SANet (a representative \textit{local transformation based} method)~\cite{SANet}. AdaIN applies mean and standard deviation to transform the features globally, and is the basis for many global style transfer methods, while SANet is the pioneering work that introduces attention mechanism to consider local feature matching.

In our context, a \textit{ticket} means a style transfer subnetwork, a \textit{winning ticket} represents a subnetwork which can match the performance of the original full style transfer model. 

In order to enable searching for winning tickets across the whole style transfer networks, we first conduct a comparative experiment to verify whether training the VGG encoder together could be comparable or even outperform the traditional training strategy, \emph{i.e.,} only training the decoder and feature transformation module. This has rarely been explored in previous studies. We evaluate the performance both quantitatively and qualitatively. Table~\ref{tab:together} and Figure~\ref{fig:together} illustrate the superiority of the strategy of training together. This finding not only benefits to obtain much sparser matching subnetworks (winning tickets), but also establishes a stronger baseline of full models. With the above discovery, we are able to further explore \textit{``Are there winning tickets in the whole style transfer models?''} 

The process of finding and verifying winning tickets uses a technique called \textit{Iterative Magnitude Pruning} (IMP), which prunes the model by alternating between network pruning and network re-training. At each iteration of this process, we obtain a sparse subnetwork (a ticket) along with its parameter initialization or rewinding. Figure~\ref{fig:Framework} gives an overview of the overall process. We fully combine the provided \textit{average style transfer test error}, visual results as well as a user study to evaluate the performance of subnetworks qualitatively and quantitatively. Furthermore, extensive experimental results demonstrate that \textit{``Style transfer models can play lottery tickets too.''}

In order to gain more insight into LTH of style transfer models, we further conduct widely verified and comparative experiments, including 1) the selection of the pruning strategy and initialization policy, 2) whether to prune feature transformation module, 3) the effect of rewinding late, 4) performance comparison with other pruning methods, and 5) LTH in other style transfer models. Through comprehensive analysis, our main findings can be summarized as follows:
\begin{itemize}
\item \textit{Training together gains a lot}: Different from the traditional training strategy \emph{i.e.,} fixing the VGG encoder, we train the encoder together to search for winning tickets in the whole network and obtain much sparser matching subnetworks. We also find that the performance of the original models can be further improved by adopting this training strategy.
\item \textit{Style transfer models can play lottery tickets too}: Using iterative magnitude pruning, we are able to identify the matching subnetworks at 89.2\% sparsity in AdaIN~\cite{AdaIN} and 73.7\% sparsity in SANet~\cite{SANet}. Moreover, these extreme winning tickets can achieve or even exceed the performance of the original full models.
\item \textit{The feature transformation module should also be pruned}: We experimentally find that, not only the autoencoder, the feature transformation module can also be pruned to get a sparser matching subnetwork without affecting the existence and quality of winning tickets.
\item \textit{Rewinding has minor impact}: Unlike~\cite{frankle2020linear, renda2020comparing}, we find that ``late rewinding'' technique does not have a notable effect on style transfer subnetworks.
\item \textit{Universal presence of LTH in style transfer models}: Besides AdaIN and SANet, we also verify LTH in other multiple style transfer models, including LST~\cite{LST}, MANet~\cite{deng2020arbitrary}, AdaAttN~\cite{liu2021adaattn}, and MCCNet~\cite{deng2021arbitrary} and obtain winning tickets with extreme sparsity of 93.1\%, 79\%, 73.7\%, 83.2\%, respectively. This indicates the great potential of LTH in style transfer model compression.
\end{itemize}

\section{Related Work}
\label{sec: related work}
\subsection{Neural Style Transfer}
Since Gatys \etal~\cite{Gatys} proposed the first CNN-based style transfer method which employs deep features from the pre-trained VGG-19 network~\cite{simonyan2014very}, we have witnessed a boom of neural style transfer methods in the past few years. Numerous research works have been conducted to improve the visual quality~\cite{style-swap, sheng2018avatar, WCT, AdaIN, SANet, kolkin2019style, LST, liu2021adaattn}, efficiency~\cite{johnson2016perceptual, li2016precomputed, radford2015unsupervised, ulyanov2016texture, jing2020dynamic} and flexibility~\cite{gatys2017controlling, risser2017stable, hu2020aesthetic, huo2021manifold} of style transfer models. Nevertheless, most of these approaches have the common problem of large model sizes due to the widespread adoption of the over-parameterized VGG-based backbone in conjunction with its corresponding feature decoder. 

Different from those efforts on improving style transfer model capability regardless of model computational complexity, we focus on making style transfer models sparser and smaller. Note that recently, Wang \etal~\cite{coust} have also attempted to train smaller style transfer models based on WCT \cite{WCT} and AdaIN \cite{AdaIN}. However, our focus is different from theirs. Specifically, Wang \etal~\cite{coust} aim to handle the ultra-resolution style transfer task via knowledge distillation. In addition, they have only compressed the encoder without compressing the decoder, hence the overall network is still large. Here, we study the over-parameterization of the whole style transfer networks from the perspective of lottery ticket hypothesis, a popular concept in deep neural network nowadays which has not been introduced into the field of style transfer yet.

\subsection{Lottery Ticket Hypothesis}
Dating back to 2018, Frankle \emph{et al.}~\cite{frankle2018lottery} firstly proposed to find winning tickets via Iterative Magnitude Pruning (IMP). The lottery ticket hypothesis (LTH) has attracted widespread attention and has been evidenced in various traditional computer vision fields, such as image classification~\cite{liu2018rethinking, wang2020picking, savarese2020winning, ma2021good}, and object detection~\cite{girish2021lottery}. Recently, the properties of LTH has also been widely studied across other fields, such as natural language processing~\cite{gale2019state, yu2019playing, prasanna2020bert, chen2020lottery}, reinforcement learning~\cite{yu2019playing}, graph neural networks~\cite{chen2021unified}, life-long learning~\cite{chen2020long}, and generative adversarial networks~\cite{chen2021gans, kalibhat2020winning, chen2021ultra}. Besides, the ``rewinding late'' rule is found by~\cite{frankle2020linear, renda2020comparing} to scale up LTH to larger networks and datasets. 

Although LTH has made pioneering progress in various deep learning fields, to our best knowledge, the research of lottery tickets hypothesis in the style transfer field remains untouched. At present, style transfer has played an important role in image and video processing areas, \emph{e.g.,} movie synthesis and photo art. Therefore, it is critical to understand the parameter redundancy in such models and make them small without sacrificing the performance.

\section{Preliminaries}
\label{sec: preliminaries}
In this section, we describe the techniques that we use to find winning tickets and the metrics to evaluate the performance of the subnetworks. We also present our setup for the empirical study.

\subsection{Original Full Networks} 
We use two representative style transfer networks as the main testbeds: AdaIN~\cite{AdaIN} and SANet~\cite{SANet}. 

AdaIN is one of the most popular global transformation based style transfer approaches. The core idea 
is to adaptively transfer the mean and standard deviation from the style feature map to the content feature map. Specifically, given a content image $I_c$ and a style image $I_s$, AdaIN first adopts the first few layers (up to $relu4\_1$) of a pre-trained VGG-19 network~\cite{simonyan2014very} to encode content features $F_c$ and style features $F_s$:
\begin{equation}
  F_c, F_s = E(I_c, I_s; \theta_E)
\end{equation}
where $E$ is the encoder with parameters $\theta_E$. Next, the AdaIN module replaces the channel-wise mean and standard deviation from one feature map to the other:
\begin{equation}
  AdaIN(F_c, F_s) = \sigma(F_s)(\frac{F_c-\mu(F_c)}{\sigma(F_c)})+\mu(F_s)
\end{equation}
where $\mu(F_c)$ ($\mu(F_s)$) calculates the mean of $F_c$ ($F_s$) and $\sigma(F_c)$ ($\sigma(F_s)$) calculates the standard deviation of $F_c$ ($F_s$). For simplicity, we assume $F_{cs}=AdaIN(F_c, F_s)$. Then, target features $F_{cs}$ are fed into the decoder $D$ to obtain the stylized image $I_t$:
\begin{equation}
\label{getresult}
  I_t = D(F_{cs}; \theta_D)
\end{equation}
where $\theta_D$ denotes the parameters in $D$. 

SANet is a local transformation based method. It is able to flexibly match the local semantically nearest style features onto the content features via a learnable style-attentional transformation module. Similar to AdaIN, SANet also utilizes the pre-trained VGG-based autoencoder architecture. The whole process of SANet can be divided into three stages. Firstly, the feature encoding stage:
\begin{equation}
  F_c^{41}, F_s^{41}, F_c^{51}, F_s^{51} = E(I_c, I_s; \theta_E)
\end{equation}
where $F_c^{41}$ ($F_s^{41}$) and $F_c^{51}$ ($F_s^{51}$) denote the corresponding layer VGG feature maps (\emph{i.e.,} $relu4\_1$ and $relu5\_1$) of content (style) images respectively. Secondly, the feature transformation stage:

\begin{equation}
  SANet(F_c, F_s) = T(F_c^{41}, F_s^{41}, F_c^{51}, F_s^{51}; \theta_T)
\end{equation}
where $T$ is the attention based feature transformation module with trainable parameters $\theta_T$ (different from the parameter-free transformation in AdaIN). Finally, the stylized output image $I_{t}$ is synthesized by feeding $F_{cs}$ into the decoder just like Eq.~(\ref{getresult}).

Actually, almost all the mainstream feed-forward style transfer methods~\cite{AdaIN, style-swap, LST, SANet, deng2020arbitrary, deng2021arbitrary, sheng2018avatar, liu2021adaattn} have similar architectures as AdaIN~\cite{AdaIN, style-swap, sheng2018avatar} or SANet~\cite{SANet, LST, deng2020arbitrary, deng2021arbitrary, liu2021adaattn}. It indicates that if LTH exists in AdaIN and SANet, it will also exist in these methods. Our experiments in Section~\ref{sec: universality} demonstrate this point.

\subsection{Subnetworks}
For a network $f$ that maps samples $x \in \mathcal{X}$ with parameters $\boldsymbol{\theta} \in \mathbb{R}^{d}$ to $f(x; \boldsymbol{\theta})$, a subnetwork is defined as $f(x; \boldsymbol{m} \odot \boldsymbol{\theta})$, where $\boldsymbol{m} \in \{0, 1\}^d$ is a binary pruning mask indicating which part of the network parameters is set to 0, with $\odot$ denoting element-wise multiplication. For any configuration $\boldsymbol{m}$, the effective parameter space of the induced network $f(x; \boldsymbol{m} \odot \boldsymbol{\theta})$ is $\{ \boldsymbol{m} \odot \boldsymbol{\theta} \ | \ \boldsymbol{\theta} \in \mathbb{R}^{d} \}$, a $\|\boldsymbol{m}\|_0$-dimensional space, hence we say that the subnetwork has $\|\boldsymbol{m}\|_0$ 
parameters instead of $d$. Specifically, for both AdaIN and SANet, two separate masks, $\boldsymbol{m_E}$ and $\boldsymbol{m_D}$, are required for the VGG based encoder and decoder. Moreover, a transformation module mask $\boldsymbol{m_T}$ is needed for SANet. Accordingly, a general subnetwork of style transfer methods consists of: a sparse encoder $E(\cdot; \boldsymbol{m_E} \odot \boldsymbol{\theta_E})$, a sparse transformation module $T(\cdot; \boldsymbol{m_T} \odot \boldsymbol{\theta_T})$ and a sparse decoder $D(\cdot; \boldsymbol{m_D} \odot \boldsymbol{\theta_D})$.

\subsection{Matching Subnetworks} 
For a network $f$ and randomly-initialized parameters $\boldsymbol{\theta^{(0)}}$, a matching subnetwork $f^*$ is given by a configuration $\boldsymbol{m} \in \{0, 1\}^d$, which is trained in isolation from $\boldsymbol{\theta^{*(0)}} = \boldsymbol{\theta^{(k)}} \odot \boldsymbol{m}$, where $\boldsymbol{\theta^{(k)}}$ is the collection of parameter values obtained by training $f$ from $\boldsymbol{\theta^{(0)}}$ for $k$ iterations. Furthermore, to be a matching subnetwork, $f^*$ should reach or even surpass the performance of a trained $f$ given the same training iterations.

\textbf{Winning Ticket.} A matching subnetwork $f^*$ is a \emph{winning ticket} if it can be trained in isolation from initialization. In other words, a winning ticket is a matching subnetwork such that $k=0$ in the definition above. Ticket search is to identify such a subnetwork, given an unpruned dense network $f$ and randomly initialized parameters $\boldsymbol{\theta^{(0)}}$. 

\begin{algorithm}[htp]  
\caption{Iterative Magnitude Pruning for Style Transfer Winning Tickets}  
\label{alg:IMP}  
\begin{algorithmic}[1]  
\REQUIRE  
Total training iteration $N$; Rewind iteration $r \ge 0$; Desired sparsity $s$
\ENSURE 
A sparse style transfer model $E(\cdot; \boldsymbol{m_E} \odot \boldsymbol{\theta_E})$, $D(\cdot; \boldsymbol{m_D} \odot \boldsymbol{\theta_D})$ and $T(\cdot; \boldsymbol{m_T} \odot \boldsymbol{\theta_T})$

\STATE Set $\boldsymbol{\theta_E}^{(r)}$, $\boldsymbol{\theta_D}^{(r)}$ and $\boldsymbol{\theta_T}^{(r)}$ as initial weights of \STATE $\ \ $ $E(\cdot)$, $D(\cdot)$ and $T(\cdot)$ respectively. 

\STATE Set $\boldsymbol{m_E} = \boldsymbol{1} \in \mathbb{R}^{\|\boldsymbol{\theta_E}^{(r)}\|_0}$, $\boldsymbol{m_D} = \boldsymbol{1} \in \mathbb{R}^{\|\boldsymbol{\theta_D}^{(r)}\|_0}$, \STATE $\ \ $ and $\boldsymbol{m_T} = \boldsymbol{1} \in \mathbb{R}^{\|\boldsymbol{\theta_T}^{(r)}\|_0}$, assume $\boldsymbol{m}=\{\boldsymbol{m_E}, \boldsymbol{m_D}, \boldsymbol{m_T}\}.$
\WHILE{the sparsity of $\boldsymbol{m} < \boldsymbol{s}$}
    \STATE Train $E(\cdot; \boldsymbol{m_E} \odot \boldsymbol{\theta_E}^{(r)})$ and $D(\cdot; \boldsymbol{m_D} \odot \boldsymbol{\theta_D}^{(r)})$ for $N$ iterations to get parameters $\boldsymbol{\theta_E}^{N}$ and $\boldsymbol{\theta_D}^{N}.$ 
    \IF{pruning the transformation module $T(\cdot)$}
        \STATE Prune $20\%$ of the parameters in $\boldsymbol{\theta_E}^{N}$, $\boldsymbol{\theta_D}^{N}$ and $\boldsymbol{\theta_T}^{N}$, calculating three mask $\boldsymbol{m_E}^{\prime}$, $\boldsymbol{m_D}^{\prime}$ and $\boldsymbol{m_T}^{\prime}.$
    \ELSE
        \STATE Prune $20\%$ of the parameters in $\boldsymbol{\theta_E}^{N}$ and  $\boldsymbol{\theta_D}^{N}$, calculating two mask $\boldsymbol{m_E}^{\prime}$ and $\boldsymbol{m_D}^{\prime}.$ $\boldsymbol{m_T}^{\prime}$ remains $\boldsymbol{1} \in \mathbb{R}^{\|\boldsymbol{\theta_T}^{(r)}\|_0}.$
    \ENDIF
        \STATE Update $\boldsymbol{m_E} = \boldsymbol{m_E}^{\prime}$, $\boldsymbol{m_D} = \boldsymbol{m_D}^{\prime}$ and $\boldsymbol{m_T} = \boldsymbol{m_T}^{\prime}.$
\ENDWHILE
\end{algorithmic}  
\end{algorithm}

\subsection{Identifying Subnetworks}
Identifying subnetworks is to find three masks $\boldsymbol{m_E}$, $\boldsymbol{m_D}$ and $\boldsymbol{m_T}$ for the encoder, decoder and transformation module, respectively. Note that $\boldsymbol{m_T}$ is not needed for AdaIN. To achieve this, we utilize the Iterative Magnitude Pruning (IMP) algorithm~\cite{han2015deep}. In particular, we determine the pruning mask $\boldsymbol{m}=(\boldsymbol{m_E}, \boldsymbol{m_D}, \boldsymbol{m_T})$ by training the full unpruned style transfer network. Then, we prune individual weights with the lowest-magnitudes throughout the network globally. In detail, the position of a remaining weight in $\boldsymbol{m}$ is marked as $1$, and the position of a pruned weight is marked as $0$. We set the weights of this subnetwork to $\boldsymbol{\theta^{(i)}}$ for a specific \emph{rewinding} step $i$ during training. For instance, to set the weights of the subnetwork to their values from the initialization, we set $\boldsymbol{\theta} = \boldsymbol{\theta^{(0)}}$. As previous work has shown, to find the smallest possible matching subnetworks, it is better to repeat this pruning process iteratively. Intuitively, we prune a certain amount (\emph{e.g.,} $20\%$) of non-zero parameters each step and retrain the network several times to reach the desired sparsity rather than pruning the network only once to meet the sparsity requirement. Algorithm~\ref{alg:IMP} presents details of the IMP procedure to find matching subnetworks. In addition, Figure~\ref{fig:Framework} also provides a flowchart to illustrate the process.

\subsection{Evaluation of Subnetworks}
To evaluate whether the subnetwork is a matching subnetwork or not, after obtaining the subnetwork $E(\cdot; \boldsymbol{m_E} \odot \boldsymbol{\theta_E})$, $D(\cdot; \boldsymbol{m_D} \odot \boldsymbol{\theta_D})$ and $T(\cdot; \boldsymbol{m_T} \odot \boldsymbol{\theta_T})$, we reset the weights to $\boldsymbol{\theta_E^{(r)}}$, $\boldsymbol{\theta_D^{(r)}}$ and $\boldsymbol{\theta_T^{(r)}}$ ($r > 0$ if rewinding strategy is used, where $r$ is the rewind iteration). We then re-train the subnetwork, and evaluate whether it can still achieve the performance as the original full network. Note that, all the test content images, as well as all the test style images have \textit{never} been seen during the training process.

\textbf{Quantitative evaluations.} 
We compare the subnetworks and the original full network via the \emph{average style transfer test error} $\mathcal{E}$ calculated from numerous stylized results based on test images.
\begin{equation}
  \mathcal{E} = \mathcal{E}_{content} + \mathcal{E}_{style}
\end{equation}

For both AdaIN and SANet models, the average style error $\mathcal{E}_{style}$ is calculated as:
\begin{equation}
\begin{split}
  \mathcal{E}_{style} = \frac{1}{N} \sum_{n=1}^{N} ( \sum_{i=1}^{L}\|\mu(\Phi_i(I_t)) - \mu(\Phi_i(I_s)) \|_2 \quad + \\ \qquad \qquad \sum_{i=1}^{L}\|\sigma(\Phi_i(I_t)) - \sigma(\Phi_i(I_s)) \|_2)
\end{split}
\end{equation}
where $N$ denotes the number of test stylized images and each $\Phi_i$ denotes the $i$-layer in VGG-19. The average content error $\mathcal{E}_c$ of AdaIN is calculated as:
\begin{equation}
\begin{split}
  \mathcal{E}_{content}^{AdaIN} = \frac{1}{N} \sum_{n=1}^{N} \| F^{41}_t - F^{41'}_c \|_2
\end{split}
\end{equation} 
where $F^{41}_t$ denotes the \emph{relu4\_1} layer feature maps of the content image and $F^{41'}_c$ denotes the content features after AdaIN transformation. Similarly, for SANet:  
\begin{equation}
\begin{split}
  \mathcal{E}_{content}^{SANet} = \frac{1}{N} \sum_{n=1}^{N} (\| F^{41}_t - F^{41'}_c \|_2 + \| F^{51}_t - F^{51'}_c \|_2 \\ + \|I_{cc}-I_c\|_2 + \|I_{ss}-I_s\|_2 + \sum_{i=1}^{L}(\|\Phi_i(I_{cc}) \\ - \Phi_i(I_{c})\|_2 + \|\Phi_i(I_{ss}) - \Phi_i(I_{s})\|_2))
\end{split}
\end{equation} 
where $I_{cc}$ (or $I_{ss}$) denotes the generated results using a common natural image (or painting) as content and style images simultaneously.

Overall, the average style transfer test error is consistent with the optimization loss function for training AdaIN and SANet networks (more details can be found in~\cite{AdaIN, SANet}). Note that in the absence of ideal evaluation metrics for assessing the style transfer models' performance, the average style transfer test error is a feasible alternative to accomplish the measurement task quantitatively. For example, combining the results of Table~\ref{tab:together} and Figure~\ref{fig:together} (or Figure~\ref{fig:IMP_OMP_RP_RT} and Figure~\ref{fig:IMP_OMP_RP_RT_IMG}), we can see that there is a strong negative correlation between visual quality and the test error, \emph{i.e.,} better performance obtained in terms of content preservation and stylization degree where there is a smaller average style transfer test error.


\textbf{Qualitative evaluations.} We also conduct a user study to qualitatively evaluate the subnetworks. Specifically, 15 content images and 20 style images are randomly picked to form 300 images pairs in total. Then we randomly sample 100 content-style pairs and synthesize stylized images using both the original full networks and the corresponding subnetworks. Results are presented side-by-side in a random order and we ask subjects to select the most visually pleasant one from three views: content preservation, stylization degree, and overall preference. We collect 2000 votes from 20 users and present the statistical results in Figure~\ref{fig:userstudy}.

\textbf{Datasets and settings.} In the training phase, we use MS-COCO dataset~\cite{lin2014microsoft} and WikiArt dataset~\cite{nichol2016painter} as our content image set and style image set, respectively. Each dataset contains roughly 80,000 training examples. Besides, we follow the same settings (\emph{e.g.,} hyperparameters, image resolution, \emph{etc.}) as described in~\cite{AdaIN, SANet} to train the original full model or its corresponding subnetworks. In the testing phase, 40 content images and 100 style images are randomly selected from the test set of the MS-COCO~\cite{lin2014microsoft} and WikiArt dataset~\cite{nichol2016painter} to calculate the average style transfer test error. All models are trained on a GeForce RTX 2080 Ti GPU.

\section{Experimental Results and Analysis}
\label{sec: experiment}

\subsection{Training Together Gains A Lot}
\label{sec:plusstrategy}

\begin{figure}[!tp]
  \centering
  \includegraphics[width=\linewidth]{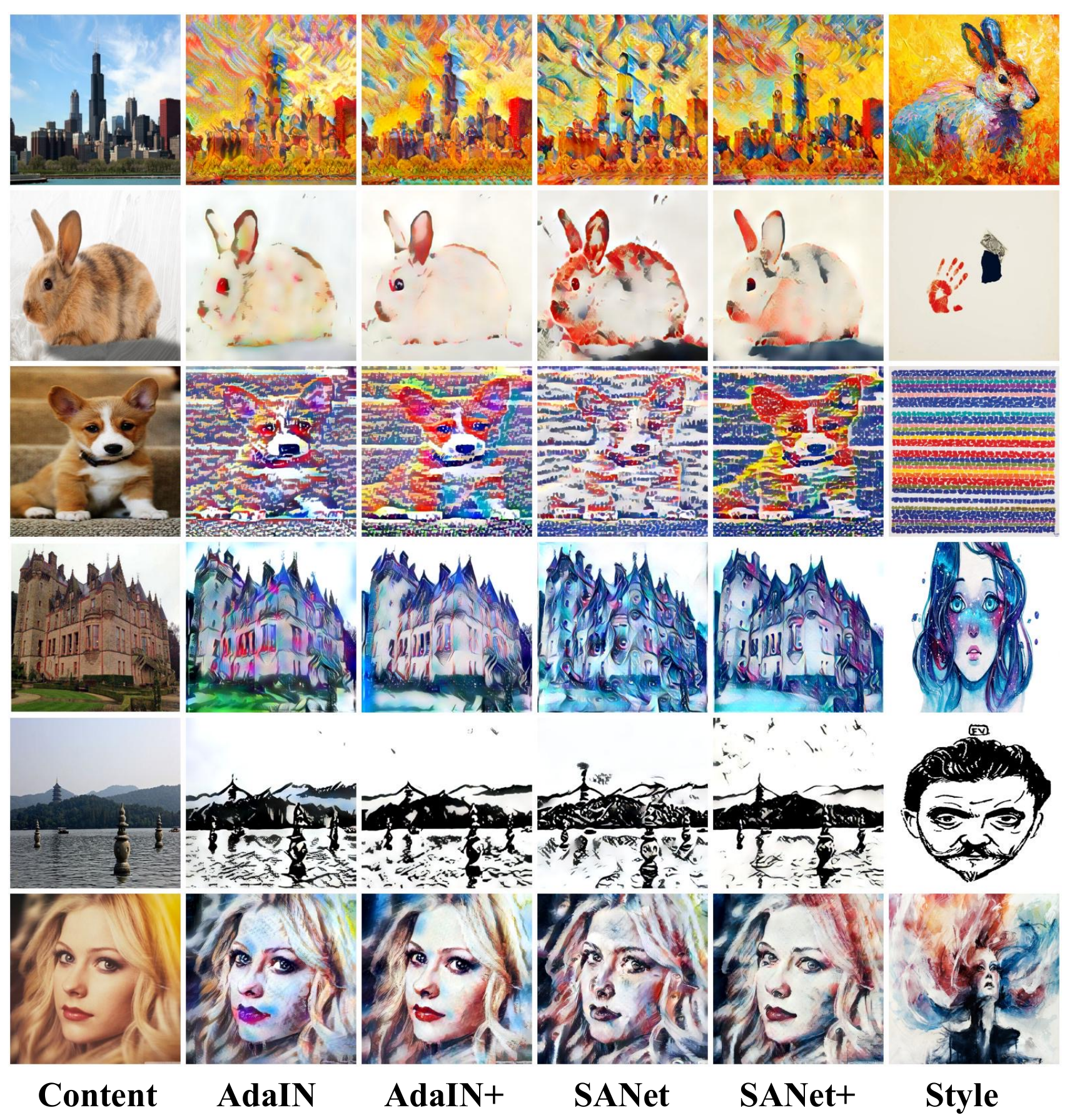}
  \caption{Image style transfer results of different model training strategies. Zoom in to have a better view.}
  \label{fig:together}
\end{figure}

Before starting the experimental verification of the LTH, we perform some preparatory experiments. As we all know, most of the existing style transfer methods adopt a common training strategy that fixes the VGG-19 encoder while only training the decoder and the feature transformation module. However, on the one hand, we are not able to search for winning tickets across the whole model while adopting the above training strategy, hence the VGG encoder is not considered during the training phase. On the other hand, it overly relies on the content and style patterns representation ability of the pre-trained VGG network which is not trained for this purpose and may not always be reliable.
Therefore, we perform experiments to compare the above training strategy and training together strategy. For simplicity, we name the training together strategy as a \textit{plus strategy}. The quantitative and qualitative results are shown in Table~\ref{tab:together} and Figure~\ref{fig:together}, respectively. 

\begin{table}[htp]
  \caption{Quantitative performance of different training strategies. AdaIN+: training encoder together based on AdaIN, SANet+: training encoder together based on SANet.}
  \label{tab:together}
  \begin{tabular}{lccc}
    \toprule
    Methods & Avg content error & Avg style error & Avg error\\
    \midrule
    AdaIN~\cite{AdaIN} & 2.284 & 6.013 & 8.297 \\
    \textbf{AdaIN+} & \textbf{2.199} & \textbf{3.486} & \textbf{5.685} \\
    \midrule
    SANet~\cite{SANet} & 10.597 & 3.841 & 14.438 \\
    \textbf{SANet+} & \textbf{5.618} & \textbf{3.009} & \textbf{8.627} \\
  \bottomrule
\end{tabular}
\end{table}

From Table~\ref{tab:together}, we can see that the plus strategy outperforms the original strategy overall in terms of content, style and total errors. In Figure~\ref{fig:together}, by comparing stylized results of the 2nd (4th) and 3rd (5th) columns, we can further determine that the plus strategy is generally better than the traditional training strategy, combined with the aspect of content preservation and stylization degree. Therefore, we claim that \textit{``Training Together Gains A Lot''}, which not only makes it possible to search for winning tickets across the whole network, but also establishes a stronger baseline of the original full model.

\begin{figure*}[htp]
  \centering
  \includegraphics[width=\linewidth]{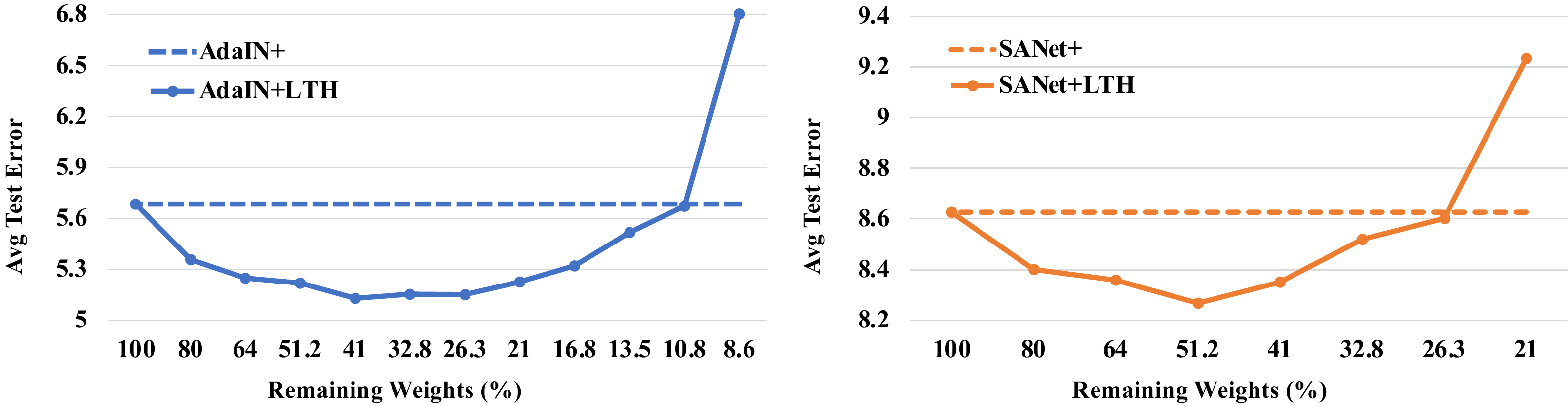}
  \caption{The average style transfer test error $\mathcal{E}$ curves of the subnetworks in AdaIN+ (left) and SANet+ (right) generated by IMP (average of three trials). The dashed line indicates the performance of full models, both trained via plus strategy as described in Section~\ref{sec:plusstrategy}. Note that the x-axis charts the \emph{percent of remaining weights}, where \emph{remaining weights (\%) $=$ 1 - sparsity (\%)}.}
  \label{fig:verifyLTH}
\end{figure*}

\begin{figure*}[htp]
  \centering
  \includegraphics[width=\linewidth]{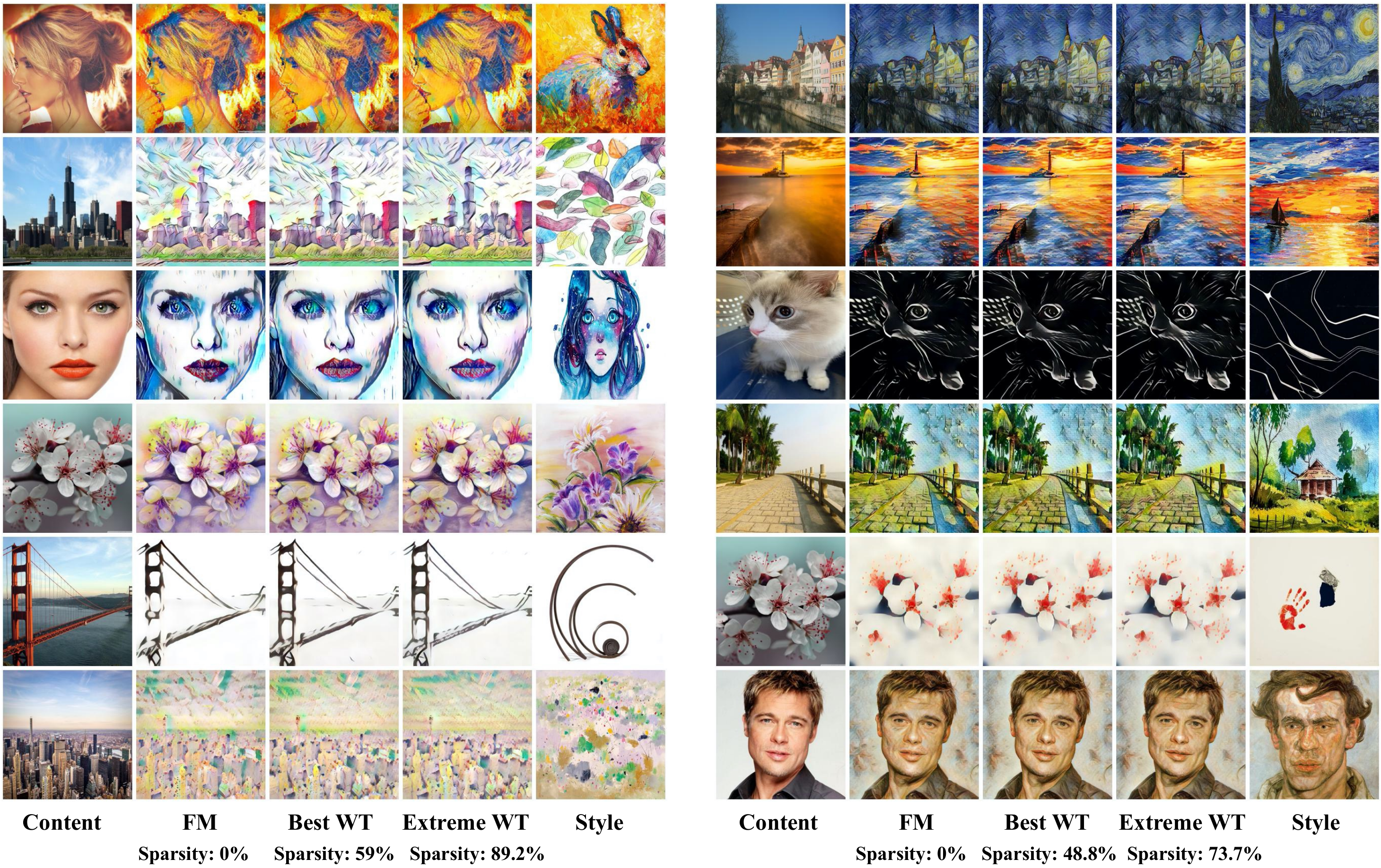}
  \caption{Image style transfer results of Winning Tickets found by IMP. Left: winning tickets of AdaIN+; Right: winning tickets of SANet+; FM: Full AdaIN+ (SANet+) model; Best WT: Winning Tickets with best performance, \emph{i.e.,} lowest test error (sparsity of 59\% for AdaIN+ and sparsity of 48.8\% for SANet+); Extreme WT: Winning Tickets with highest sparsity (89.2\% for AdaIN+ and 73.7\% for SANet+). Zoom in to have a better view.}
  \label{fig:LTHCMP_Merge}
\end{figure*}

\subsection{Style Transfer Models Can Play Lottery Tickets Too}

To prove this point, we follow the same procedure in~\cite{frankle2018lottery}. We first conduct experiments on AdaIN by pruning the VGG encoder and decoder with the following steps: 1) Run IMP to obtain the sparsity pattern $ \{\boldsymbol{m_E}^{\prime}, \boldsymbol{m_D}^{\prime}\}$, with $s_i\%$ sparsity; 2) Initialize the resulting subnetwork to $\boldsymbol{\theta_E}^{(0)}$ and $\boldsymbol{\theta_D}^{(0)}$. This produces a subnetwork $\{E(\cdot; \boldsymbol{m_E}^{\prime} \odot \boldsymbol{\theta_E}^{(0)}), D(\cdot; \boldsymbol{m_D}^{\prime} \odot \boldsymbol{\theta_D}^{(0)})\}$; 3) Train this subnetwork again to evaluate whether it is a winning ticket. In detail, we set $s_i\% = (1-0.8^i) \times 100\%$, which we use for all the experiments that involve iteratively pruning hereinafter. For SANet, besides $\boldsymbol{m_E}^{\prime}$ and $\boldsymbol{m_D}^{\prime}$, $\boldsymbol{m_T}^{\prime}$ is also required to produce the subnetwork $\{E(\cdot; \boldsymbol{m_E}^{\prime} \odot \boldsymbol{\theta_E}^{(0)}), \{T(\cdot; \boldsymbol{m_T}^{\prime} \odot \boldsymbol{\theta_T}^{(0)}), D(\cdot; \boldsymbol{m_D}^{\prime} \odot \boldsymbol{\theta_D}^{(0)})\}$. The overall process is similar to AdaIN. More specific details can be found in Algorithm~\ref{alg:IMP}.

Figure~\ref{fig:verifyLTH} shows the quantitative performance of the subnetworks generated by IMP with different sparsity levels. As can be seen, we are still able to find the winning tickets by iteratively pruning the entire networks to very high sparsity, around $90\%$ in AdaIN+, and around $74\%$ in SANet+, where the test errors of these subnetworks are comparable or even less than the full networks. Importantly, the graphs also suggest that the subnetworks at some sparsity levels perform statistically significantly better than the full networks.

\begin{figure}[htp]
  \centering
  \includegraphics[width=\linewidth]{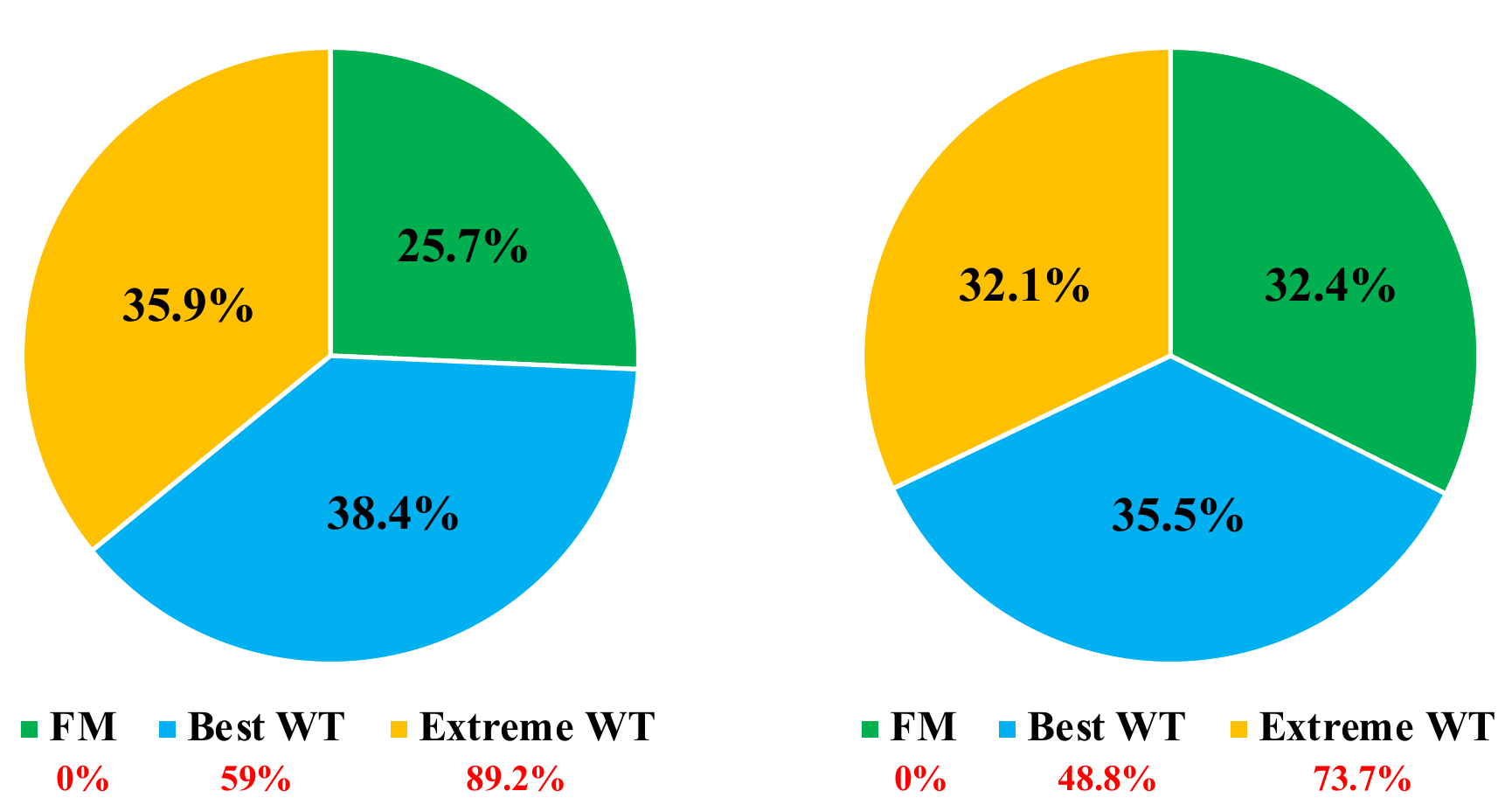}
  \caption{User study of Winning Tickets found by IMP. Left: winning tickets of AdaIN+; Right: winning tickets of SANet+. The percentage in red shown underneath each method indicates the sparsity of corresponding network.}
  \label{fig:userstudy}
\end{figure}

To give a more intuitive impression, we visualize the stylized results in Figures~\ref{fig:LTHCMP_Merge}. For AdaIN+, we can see that the extremely sparse subnetwork performs very well even with only 10.8\% parameters of the full network. Surprisingly, the found winning tickets in AdaIN+ can even yield more pleasant stylized results with fewer artifacts and better content preservation in detail, \emph{e.g.,} less artifacts at the edges of the buildings (row 2), and in the sky (row 6), more complete bridge structure (row 5). We speculate that this phenomenon may be explained since a model with fewer parameters tends to be less overfitting. Meanwhile, from the stylized results generated by winning tickets found in SANet+, we can also observe that in the case of high sparsity, \emph{i.e.,} 73.7\%, the matching subnetwork can still achieve comparable performance to the original full model. 

Moreover, we conduct a user study to evaluate the quality of the winning tickets more objectively. The statistical results are presented in Figure~\ref{fig:userstudy}. We can see that no matter AdaIN+ or SANet+, the found winning tickets can achieve the performance that is comparable or even outperforms the full models, \emph{i.e.,} 38.4\% \emph{vs.} 25.7\% for AdaIN+ and 35.5\% \emph{vs.} 32.4\% for SANet+. Besides AdaIN and SANet, the existence of the LTH can also be confirmed in many other style transfer models as shown in Section~\ref{sec: universality}. In summary, through the comparison and analysis above, we can conclude that \textit{``Style Transfer Models Can Play Lottery Tickets Too"}.

\subsection{Feature Transformation Module Should Also be Pruned}

Actually, for a range of feature transformation based style transfer models with learnable parameters~\cite{SANet, LST, deng2020arbitrary, deng2021arbitrary, liu2021adaattn}, the intermediate transformation modules also contain a large number of parameters, sometimes even more than the total parameters of the decoder. Taking SANet as an example, the parameter sizes of the encoder, transformation module and decoder are 49.38MB, 17.02MB and 13.37MB respectively, which means that even all parameters of autoencoder are pruned, only up to 78.7\% sparsity can be achieved. Therefore, a natural question that came to our mind is: \textit{Can we prune the feature transformation module to get a much sparser style transfer network without compromising performance?}

\begin{figure}[htp]
  \centering
  \includegraphics[width=\linewidth]{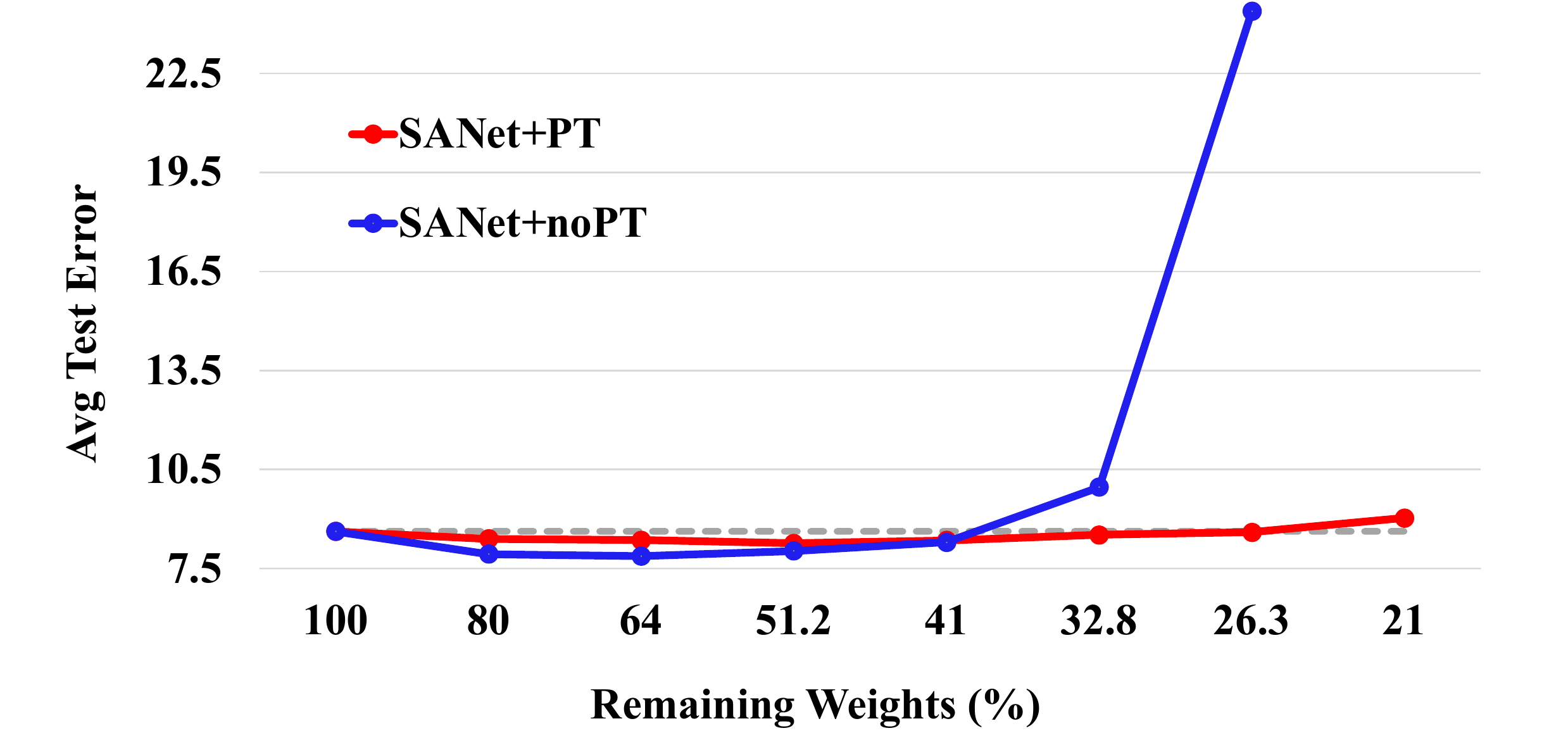}
  \caption{Results of whether to prune the feature transformation module of SANet. PT: Pruning Transformation module; noPT: not Pruning Transformation module. The dashed line indicates the performance of the full SANet+ model.}
  \label{fig:PruneT1}
\end{figure}

\begin{figure}[htp]
  \centering
  \includegraphics[width=\linewidth]{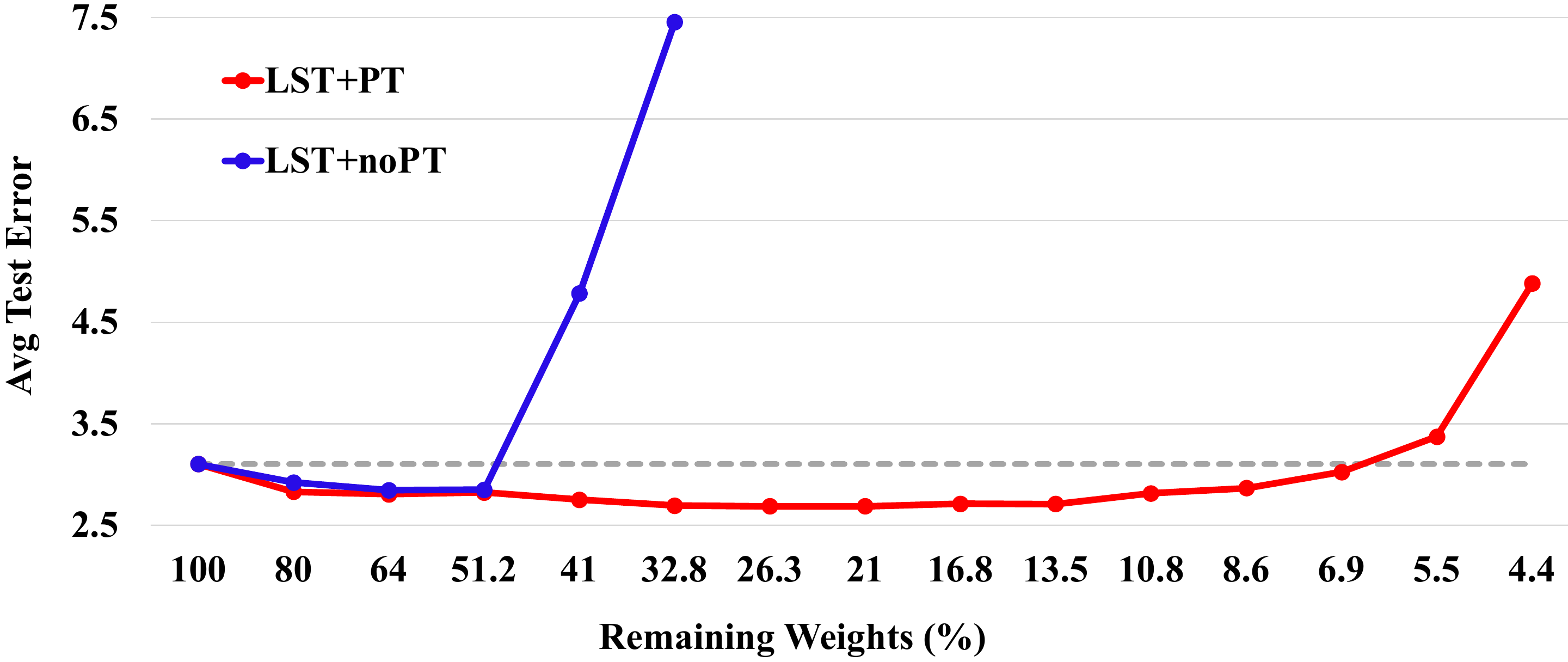}
  \caption{Results of whether to prune the feature transformation module of LST. The dashed line indicates the performance of the full LST+ model.}
  \label{fig:PruneT2}
\end{figure}

To answer this question, we compare two different iterative pruning settings for both SANet~\cite{SANet} and LST~\cite{LST}: 1) Prune the autoencoder only (noPT); 2) Prune both autoencoder and the transformation module (PT). All parts of the network are reset to the same random initialization $\boldsymbol{\theta}^{(0)}$ after the masks are obtained.

The experimental results are reported in Figure~\ref{fig:PruneT1} and Figure~\ref{fig:PruneT2}, respectively. Both graphs suggest that the two settings share similar patterns when subnetworks are not too sparse (\emph{i.e.,} 0\%--50\% of sparsity). However, as the total number of parameters of the model is further reduced to 30\% or less, PT strategy shows enormous advantage compared to noPT strategy. The explanation is probably that adopting the noPT strategy, the parameters in the encoder and decoder will be severely reduced to get higher sparsity overall, hence damaging the quality of the subnetworks. While the PT strategy can achieve the overall trade-off of the whole model parameters and still perform well with extremely high sparsity, as the transformation module is also pruned.

\subsection{IMP Winning Tickets Are Sparser than OMP, Random Pruning, and Random Tickets}

Previous work describes winning tickets as a ``combination of weights and connections capable of learning''~\cite{frankle2018lottery}, which means both the specific pruned weights and the specific initialization are necessary 
for a winning ticket to achieve this performance. 
To extend such a statement in the context of style transfer models, we compare IMP with several other benchmarks, 
one-shot magnitude pruning (OMP), random pruning (RP), and random tickets (RT). 
Specifically, we train a subnetwork $\{E(\cdot; \boldsymbol{m_E^{OMP}} \odot \boldsymbol{\theta_E}^{(0)}), D(\cdot; \boldsymbol{m_D^{OMP}} \odot \boldsymbol{\theta_D}^{(0)})\}$ with a one-hot magnitude pruning mask $\boldsymbol{m^{OMP}}$ (which evaluates the importance of the iterative pruning strategy), 
a subnetwork 
$\{E(\cdot; \boldsymbol{m_E^{RP}} \odot \boldsymbol{\theta_E}^{(0)}), D(\cdot; \boldsymbol{m_D^{RP}} \odot \boldsymbol{\theta_D}^{(0)})\}$ 
with a random pruning mask $\boldsymbol{m^{RP}}$ (which evaluates the importance of the pruning masks) and a subnetwork $\{E(\cdot; \boldsymbol{m_E^{IMP}} \odot \boldsymbol{\theta_E}^{(0)^{\prime}}), D(\cdot; \boldsymbol{m_D^{IMP}} \odot \boldsymbol{\theta_D}^{(0)^{\prime}})\}$ with a random initialization $\boldsymbol{\theta}^{\prime}$ (which evaluates the importance of the pre-trained initialization) based on AdaIN+ to see if IMP can obtain the best performance.

\begin{table}[htp]
  \caption{Results of the best subnetworks and the extreme sparsity of matching networks found by Iterative Magnitude Pruning, One-hot Magnitude-based Pruning, Random Pruning and Random Tickets. $\mathcal{E}_{Best}$: The minimal test error of all subnetworks. $\mathcal{S}_{Extreme}$: Extreme sparsity where matching subnetworks exist.}
  \label{tab:IMP_OMP_RP_RT}
  \begin{tabular}{lcc}
    \toprule
    Methods & $\mathcal{E}_{Best}$(Sparsity) & $\mathcal{S}_{Extreme}$ \\
    \midrule
    Full Model & 5.685(0.0\%) & -- \\
    \midrule
    \textbf{IMP} & \textbf{5.134(59.0\%)} & \textbf{89.2\%} \\
    OMP & 5.255(59.0\%) & 59.0\% \\
    Random Pruning & 5.413(20.0\%) & 30.0\% \\
    Random Tickets & -- & 0.0\% \\
  \bottomrule
\end{tabular}
\end{table}

\begin{figure}[htp]
  \centering
  \includegraphics[width=\linewidth]{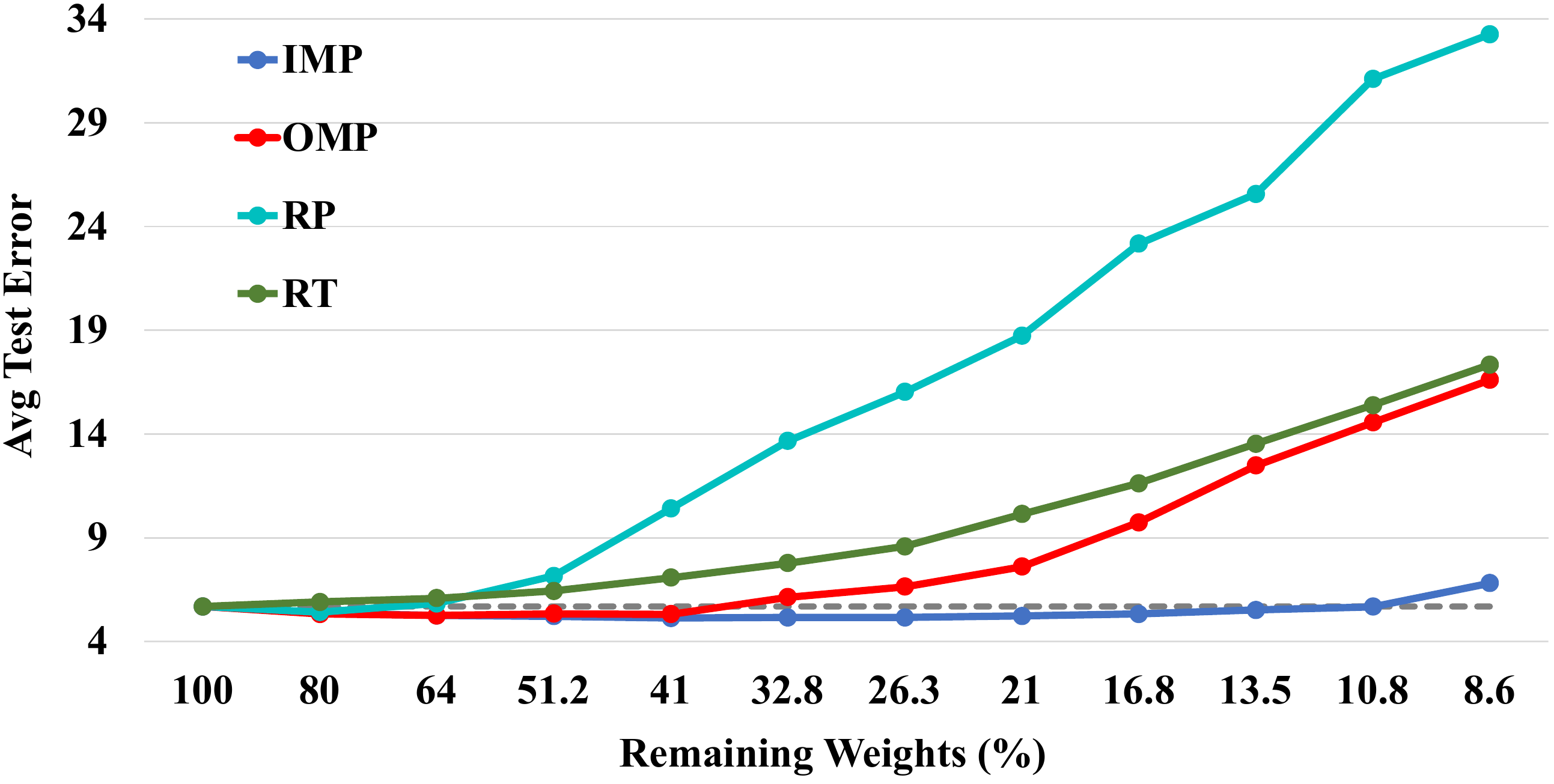}
  \caption{Test error $\mathcal{E}$ curves of subnetworks generated by different pruning settings. IMP: iterative magnitude pruning; OMP: one-hot magnitude
pruning; RP: iterative \textit{random} pruning; RT: iterative pruning but with the weights reset \textit{randomly}. The dashed line indicates the performance of the full AdaIN+ model.}
  \label{fig:IMP_OMP_RP_RT}
\end{figure}

\begin{figure}[htp]
  \centering
  \includegraphics[width=\linewidth]{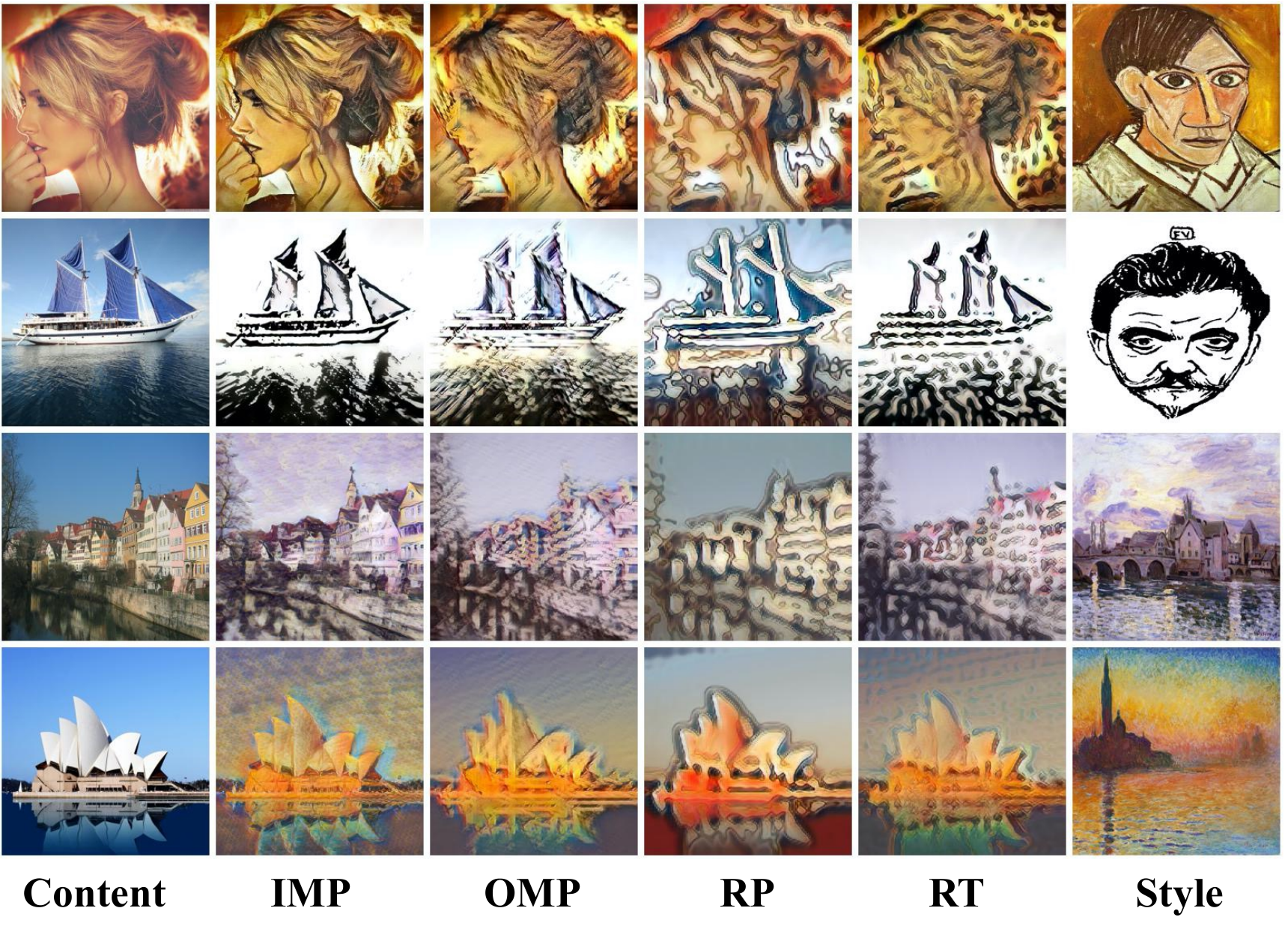}
  \caption{Image style transfer results of IMP, OMP, RP and RT with \textit{sparsity of 89.2\%}. Zoom in to have a better view.}
  \label{fig:IMP_OMP_RP_RT_IMG}
\end{figure}

 As shown in Table~\ref{tab:IMP_OMP_RP_RT} and Figure~\ref{fig:IMP_OMP_RP_RT}, the extreme winning ticket found by IMP is significantly sparser than that found by other pruning settings, and the minimal test error is smaller as well, which confirms the superiority of iterative pruning. Moreover, when adopting random tickets, we cannot obtain any matching subnetworks. This observation defends that specific initialization is essential for finding winning tickets. In Figure~\ref{fig:IMP_OMP_RP_RT_IMG}, we also show the visual performance of subnetworks with sparsity of 89.2\% obtained by different pruning strategies. We observe that all other subnetworks have poor performance in style transfer with such a high sparsity, except for the IMP winning ticket.

\subsection{Whether Rewinding Improves the Performance}

In previous paragraphs, we demonstrate that we are able to find winning tickets in both AdaIN and SANet at non-trivial sparsities (\emph{i.e.}, sparsities where random pruning cannot find winning tickets). Considering that the work~\cite{renda2020comparing} shows the rewinding paradigm is necessary to identify winning tickets for large networks, we would like to examine whether rewinding is helpful in the context of style transfer models. We perform experiments at different rewinding ratios. Specifically, after IMP training, we obtain the masks. Then, instead of resetting the weights to $\boldsymbol{\theta^{(0)}}$, we rewind the weights to $\boldsymbol{\theta^{(i)}}$, \emph{i.e.}, the weights after $i$ steps of training. The rewinding ratio = $i/N$, where $N$ is the total training iteration. The results are shown in Table~\ref{tab:Rewind}, we can see that rewinding does not have a notable effect. In particular, subnetworks trained at different rewinding ratios have the similar highest sparsity and best performance.
\begin{table}[htp]
  \caption{Rewinding results of the best subnetworks and the extreme sparsity of matching networks found by IMP.}
  \label{tab:Rewind}
  \begin{tabular}{l|cc|cc}
    \hline
    \multirow{2}{*}{\shortstack{Rewinding\\ratios}} & \multicolumn{2}{c}{AdaIN} & \multicolumn{2}{c}{SANet} \\
    \cline{2-5} & $\mathcal{E}_{Best}$ & $\mathcal{S}_{Extreme}$ & $\mathcal{E}_{Best}$ & $\mathcal{S}_{Extreme}$ \\
     \hline
    Rewind 0\% & 5.134 & 89.2\% & 8.268 & 73.7\%\\
    Rewind 10\% & 5.114 & 89.2\% & 8.166 & 73.7\%\\
    Rewind 20\% & 5.097 & 89.2\% & 8.274 & 73.7\%\\
    Rewind 30\% & 5.052 & 89.2\% & 8.012 & 73.7\%\\
    Rewind 40\% & 5.083 & 89.2\% & 8.249 & 73.7\%\\
   \hline
\end{tabular}
\end{table}

\subsection{IMP Winning Tickets vs. Other Pruning Methods}

\begin{figure}[htp]
  \centering
  \includegraphics[width=\linewidth]{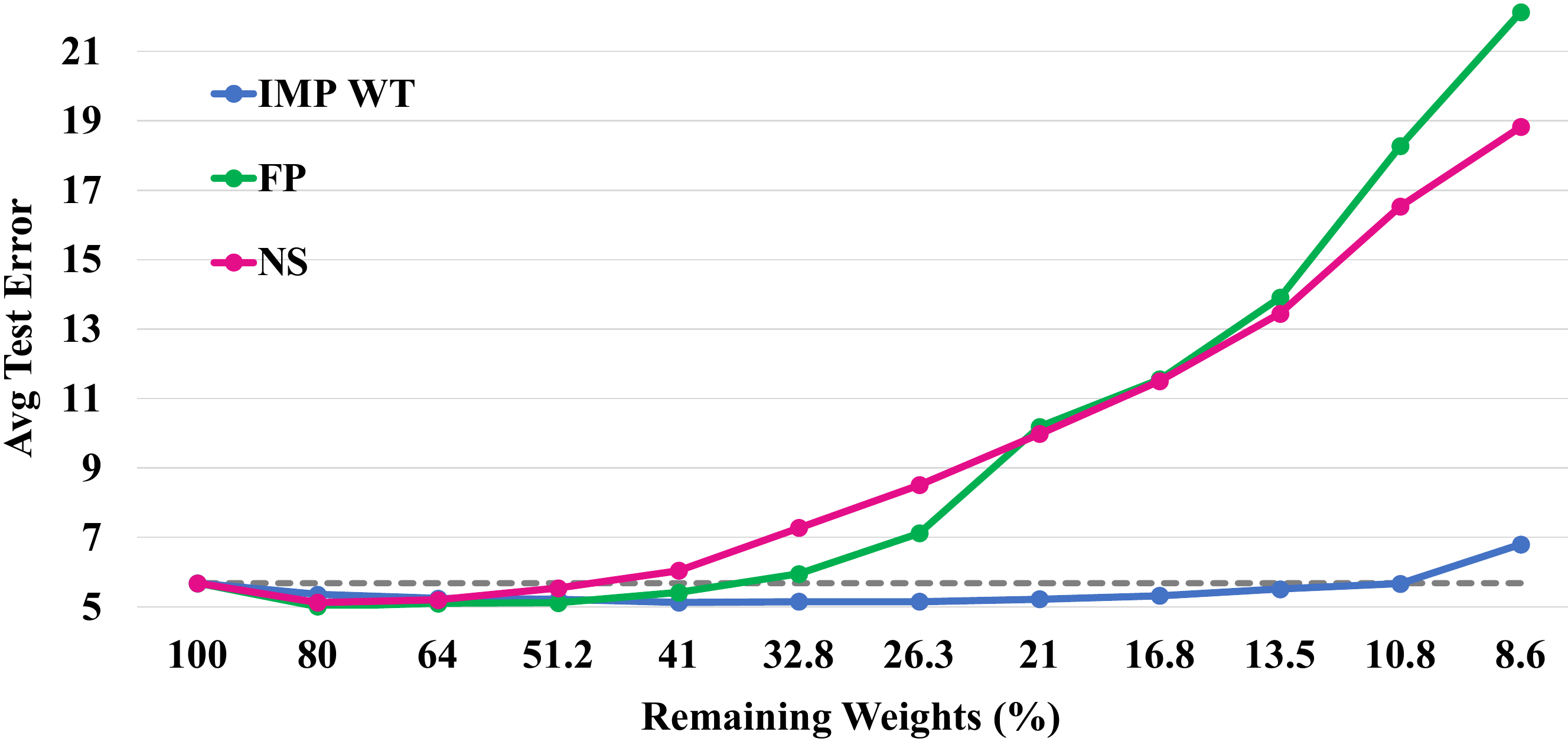}
  \caption{Results of different pruning methods. IMP WT: IMP Winning Tickets; FP: Finetuning based magnitude Pruning; NS: Network Slimming. The dashed line indicates the performance of the original AdaIN+ model.}
  \label{fig:IMP_NS_Finetune}
\end{figure}

We further conduct experiments to compare the IMP Winning Tickets with other mainstream pruning methods, \emph{i.e.,} the structured channel pruning method~\cite{liu2017learning} (network slimming) and finetuning based magnitude pruning method~\cite{han2015learning}. Results are shown in Figure~\ref{fig:IMP_NS_Finetune}. We can find that the IMP winning tickets are significantly better than subnetworks found by network slimming and finetuning based magnitude pruning.

\subsection{Experiments on Other Style Transfer Models}
\label{sec: universality}

To verify the universal presence of the lottery ticket hypothesis in diverse style transfer models, we also conduct 
experiments on LST~\cite{LST}, MANet~\cite{deng2020arbitrary}, AdaAttN~\cite{liu2021adaattn} and MCCNet~\cite{deng2021arbitrary}. Table~\ref{tab:MoreST} shows that LTH can be generalized to various style transfer models despite the different extreme sparsities. Besides, we are surprised to find that the smallest matching subnetwork of LST has only 6.9\% of the parameters of the original full model.
\begin{table}[htp]
  \caption{Results on other style transfer models. $\mathcal{E}_{Full}$: The test error of the full model. $\mathcal{E}_{Best}$: The minimal test error of all subnetworks. $\mathcal{S}_{Extreme}$: Extreme sparsity where matching subnetworks exist. ($\cdot$\%) denotes the sparsity of the corresponding network.} 
  \label{tab:MoreST}
  \begin{tabular}{cccc}
    \toprule
    Model & $\mathcal{E}_{Full}$ & $\mathcal{E}_{Best}$ & $\mathcal{S}_{Extreme}$ \\
    \midrule
    LST~\cite{LST} & 3.103(0\%) & 2.686(79.0\%) & \textbf{93.1\%} \\
    MANet~\cite{deng2020arbitrary} & 17.176(0\%) & 11.227(20.0\%) & \textbf{79.0\%} \\
    AdaAttN~\cite{liu2021adaattn} & 24.467(0\%) & 22.873(48.8\%) & \textbf{73.7\%} \\
    MCCNet~\cite{deng2021arbitrary} & 7.875(0\%) & 7.574(59.0\%) & \textbf{83.2\%} \\
  \bottomrule
\end{tabular}
\end{table}

\section{Conclusion}
\label{sec: conclusion}
In this paper, the LTH has been extended to the style transfer field for the first time. Through extensive experiments and comprehensive analysis, we verify the existence of winning tickets in a range of style transfer models. In future work, we plan to examine the speedup results on a hardware platform that is friendly to unstructured pruning. For instance, XNNPACK~\cite{elsen2020fast} has shown significant speedups over dense baselines on smartphone processors at 70\%-90\% unstructured sparsity. Besides, we will further explore the application of LTH in more diverse style transfer scenarios, \emph{e.g.,} video synthesis, caricature generation, \emph{etc.}


\bibliographystyle{ACM-Reference-Format}
\bibliography{main}


\end{sloppypar}
\end{document}